\documentclass[10pt,twocolumn,letterpaper]{article}

\usepackage{cvpr}              

\usepackage{graphicx}
\usepackage{amsmath}
\usepackage{amssymb}
\usepackage{booktabs}

\usepackage[pagebackref,breaklinks,colorlinks]{hyperref}

\usepackage[capitalize]{cleveref}
\crefname{section}{Sec.}{Secs.}
\Crefname{section}{Section}{Sections}
\Crefname{table}{Table}{Tables}
\crefname{table}{Tab.}{Tabs.}

\def\cvprPaperID{27} 
\def\confName{CVPR}
\def\confYear{2022}

\begin{document}


\title{Object State Change Classification in Egocentric Videos\\ using the Divided Space-Time Attention Mechanism}

\author{Md Mohaiminul Islam\\
UNC Chapel Hill\\
{\tt\small mmiemon@cs.unc.edu}
\and
Gedas Bertasius\\
UNC Chapel Hill\\
{\tt\small gedas@cs.unc.edu}
}
\maketitle

\begin{abstract}
   This report describes our submission called ``TarHeels" for the Ego4D: Object State Change Classification Challenge. We use a transformer-based video recognition model and leverage the Divided Space-Time Attention mechanism~\cite{gberta_2021_ICML} for classifying object state change in egocentric videos. Our submission achieves the second-best performance in the challenge. Furthermore, we perform an ablation study to show that identifying object state change in egocentric videos requires temporal modeling ability. Lastly, we present several positive and negative examples to visualize our model's predictions. The code is publicly available. \footnote{\url{https://github.com/md-mohaiminul/ObjectStateChange}} 
\end{abstract}

\section{Introduction}

There has been rapid progress in the field of video classification in recent years~\cite{carreira2017quo, feichtenhofer2019slowfast, karpathy2014large, gberta_2021_ICML, arnab2021vivit, liu2021video}. However, most existing datasets contain short videos from a third-person view. Moreover, many video recognition datasets are collected by a human in controlled environments. As a result, most of these datasets cannot be applied to many applications, for example, robotics and augmented reality, where we need to process videos from a first-person or egocentric point of view. Designing a video recognition model for egocentric videos can be difficult; because egocentric videos often contain many challenges such as camera movement, motion blur, object occlusion, etc.

In this challenge, we need to recognize how the camera wearer changes the state of an object by using or manipulating it in an egocentric video. One of the main challenges of this task is the generalization, i.e, the same object state change can be achieved in several ways. For example, a piece of wood can be cut into half in several ways, such as various tools, force, speed, grasps, etc., nevertheless, all should be recognized as the same object state change. Figure~\ref{fig:example} shows several example videos of this task. 

For this challenge, we utilize a transformer-based video recognition model. First, we divide each frame of video into non-overlapping patches. Then we linearly project each patch into an embedding and augment it with positional information. Finally, we feed the sequence of patch-level embeddings into a transformer model. We leverage the divided space-time attention (TimeSformer~\cite{gberta_2021_ICML}) mechanism for the transformer architecture, where temporal attention and spatial attention are applied separately within each block of the network. TimeSformer achieves state-of-the-art performance in several video classification benchmarks; however, most of these datasets contain videos of third-person view. In this work, we adopt TimeSformer for video recognition in egocentric videos, particularly for the object state change classification.

Our submission named ``TarHeels" achieves the second-best performance in the Ego4D: Object State Change Classification Challenge~\cite{grauman2021ego4d}. Moreover, we perform an ablation study with different attention mechanisms and show that identifying object state change in egocentric videos requires temporal modeling ability. 

\section{Our Approach}

We follow the TimeSformer model proposed by \cite{gberta_2021_ICML}. Our model takes a video clip $X \in \mathbb{R}^{H\times W\times 3\times F}$ as input, where $H$ is the height, $W$ is the width, and $F$ is the number of frames. Then we divide each frame into $N$ non-overlapping patches of size $P\times P$, where $N=HW/P^2$. This is equivalent to applying a convolution with equal kernel size and stride of $P\times P$. Then a linear layer is applied to project each patch to a latent dimension of $D$, and a positional embedding $E\in\mathbb{R}^{N\times D}$ is added to each patch embedding.

The resulting sequence of patch embedding is passed through a transformer encoder which is a stack of $L$ transformer blocks. Each transformer block contains a multi-headed attention (MHA) and a multi-layer perceptron (MLP) block. Layer normalization (LN) is applied before each block, and a skip connection layer is added after each block.

We follow the divided space-time attention schemes, which decomposes the self-attention operation over the temporal dimension and spatial dimension. First temporal attention is computed, where the self-attention weights $\boldsymbol{\alpha}^{(l,a)}_{(p,t)}$ of later $l$ and attention head $a$ for a patch at spatial location $p$ and temporal location $t$ is computed by the following equation.

\begin{equation}
\boldsymbol{\alpha}^{(l,a)time}_{(p,t)} = SM \left( \frac{{\boldsymbol{q}^{(l,a)}_{(p,t)}}^T}{\sqrt{D_h}} \cdot \left[ \boldsymbol{k}^{(l,a)}_{(0,0)}\left\{ \boldsymbol{k}^{(l,a)}_{(p,t')}\right\}_{t'=1,\dots,F}\right]\right)
\end{equation}

Here, SM denotes the softmax activation, and $D_h$ is the dimensionality of the attention heard, where, $D_h = D/A$, A being the number of attention heads. Similarly, spatial attention is computed by the following equation.

\begin{equation}
\boldsymbol{\alpha}^{(l,a)space}_{(p,t)} = SM \left( \frac{{\boldsymbol{q}^{(l,a)}_{(p,t)}}^T}{\sqrt{D_h}} \cdot \left[ \boldsymbol{k}^{(l,a)}_{(0,0)}\left\{ \boldsymbol{k}^{(l,a)}_{(p',t)}\right\}_{p'=1,\dots,N}\right]\right)
\end{equation}

\begin{figure}[t]
    \centering
    \includegraphics[width=0.5\textwidth]{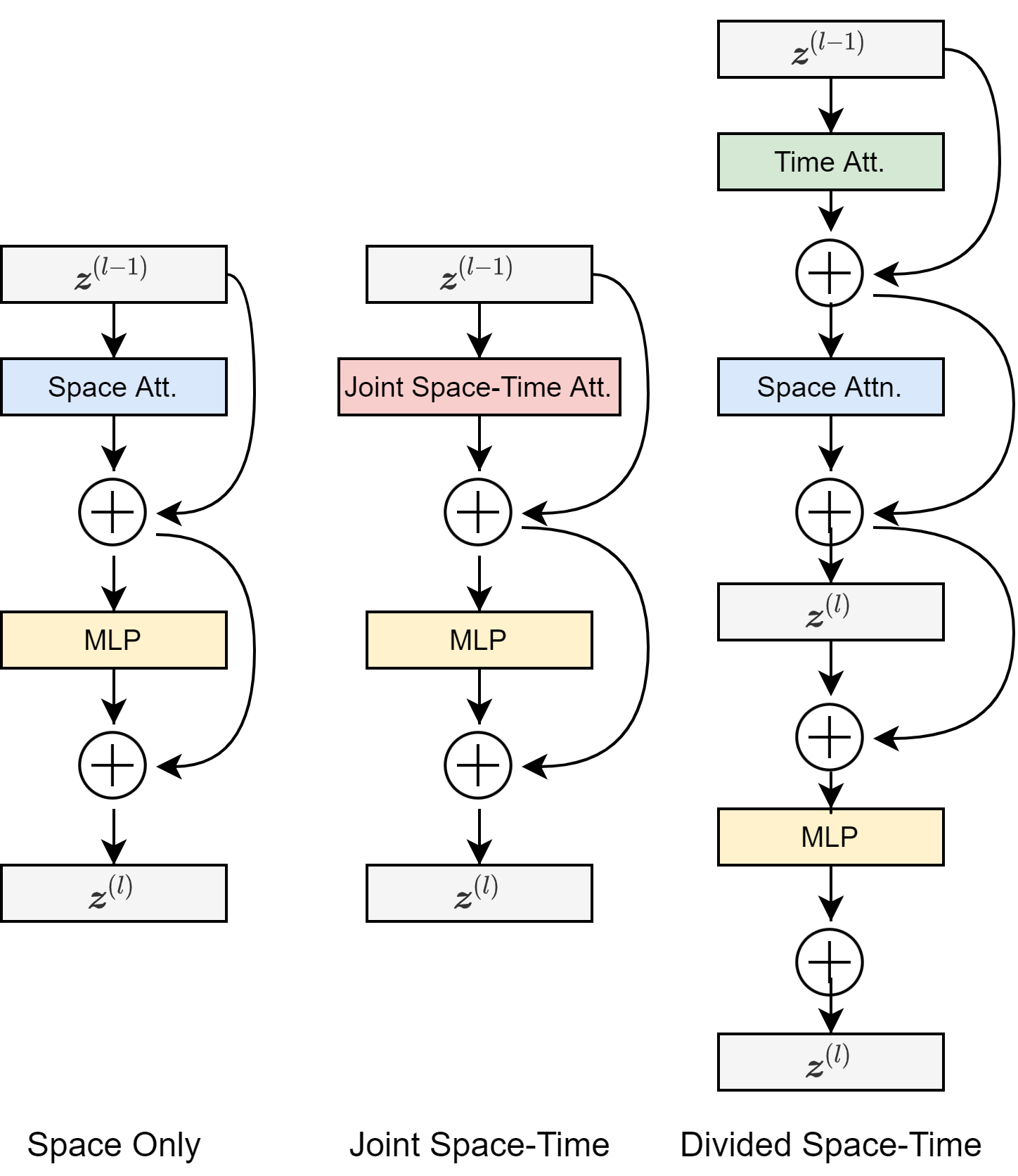}
    \caption{This figure visualizes Space Only, Joint Space-Time, and Divided Space-Time attention blocks. We use the Divided Space-Time attention mechanism for our model.}
    \label{fig:model}
\end{figure}

We also experiment with two other attention schemes, i.e, Space-Only attention and Joint Space-Time Attention. In the Space-Only attention mechanism, attention weights are computed only considering spatial tokens within the same frame. Whereas, in the Joint Space-Time attention scheme, attention weights are calculated considering all spatial and temporal tokens of the video clip. Note that, computing attention over one dimension (e.g., spatial or temporal) significantly reduces the computational cost. In Figure~\ref{fig:model}, we visualize Space-Only, Joint Space-Time, and Divided Space-Time attention blocks. 

\section{Implementation Details} 

\noindent \textbf{Model Architecture.} We follow TimeSformer-base architecture, which operates on a video of dimension $8\times224\times224$. We divide each video frame using a patch size of $16\times16$. We use ViT-base~\cite{dosovitskiy2020image} as the transformer backbone, which has $12$ self-attention layers with $12$ heads on each layer. It uses a hidden dimension $768$, and MLP dimension $3072$. The transformer backbone is pre-trained on the Imagenet-21k dataset~\cite{deng2009imagenet}. 

\noindent \textbf{Training.} We resize each video to a height and width of $256\times320$ and use a sampling rate of 3/2 fps. During training, we sample $8$ consecutive frames from a video and randomly sample a crop of the size of $224 \times224$ from each frame. Then we feed the $8$ frames video clip to our transformer model. We train our model for 15 epochs with an initial learning rate of $0.005$ and divide the learning rate by $10$ at epochs $11$, and $14$. We train our model using $8$ NVIDIA RTX A6000 GPUs using a batch size of $16$. We use the SGD optimizer with a momentum of 0.9 and weight decay of $0.0001$.

\noindent \textbf{Inference.} During inference, we randomly sample $3$ clips of $8$ frames from a video. For each temporal clip, we construct $3$ clips by randomly cropping video frames to dimensions $224\times224$. Thus, we construct $9$ clips for each video, and the final prediction is obtained by averaging the softmax scores of these 9 predictions. 

\section{Experimental Results}

\noindent \textbf{Leader Board Results.} In Table~\ref{results_main}, we present leaderboard results on the unseen test set of Ego4D: Object State Change Classification Challenge~\cite{grauman2021ego4d}. The performance is evaluated by classification accuracy. Our challenge entry named ``TarHeels" achieves second-best performance among all challenge entries

\begin{table}[t]
\begin{center}
 \begin{tabular}{c c} 
 \toprule
Method & Accuracy\\
\midrule
Dejie & 48\\
Host\_9975\_Team & 68\\
Uniandes & 68\\
TarHeels (ours) & \underline{72}\\
Egocentric VLP & \textbf{74}\\
\bottomrule
\end{tabular}
\end{center}
\caption{Leaderboard results on the unseen test set of Ego4D: Object State Change Classification Challenge~\cite{grauman2021ego4d}. The performance is evaluated by classification accuracy. Our challenge entry, named ``TarHeels", achieves second-best performance among all challenge entries.\vspace{-0cm}}
\label{results_main}
\end{table}

\noindent \textbf{Ablation Studies.} We ablate different self-attention schemes and analyze the importance of the temporal modeling for the Object State Change Classification. Particularly, we train our model 5 epochs using three self-attention mechanisms (Space-only, Joint Space-Time, Divided Space-Time) and present the performance of the validation set in Table~\ref{ablation}. First of all, we observe that Space-only attention achieves the worst performance which indicates the importance of temporal modeling for identifying object state change. Secondly, the divided Space-Time attention mechanism leads to the best accuracy showing the effectiveness of this method for this particular task.

\begin{table}[t]
\begin{center}
 \begin{tabular}{c c} 
 \toprule
Attention Scheme & Accuracy\\
\midrule
Space-only & 69.06\\
Joint Space-Time & 70.42\\
Divided Space-Time & \textbf{70.81}\\
\bottomrule
\end{tabular}
\end{center}
\caption{The performance of different self-attention schemes on the validation set after 5 epochs. We observe that space-only attention achieves the worst performance which indicates the significance of the temporal modeling for this task. The divided Space-Time attention mechanism leads to the best accuracy. \vspace{-0cm}}
\label{ablation}
\end{table}

\section{Qualitative Results}

\begin{figure*}[t]
    \centering
    \includegraphics[width=1\textwidth]{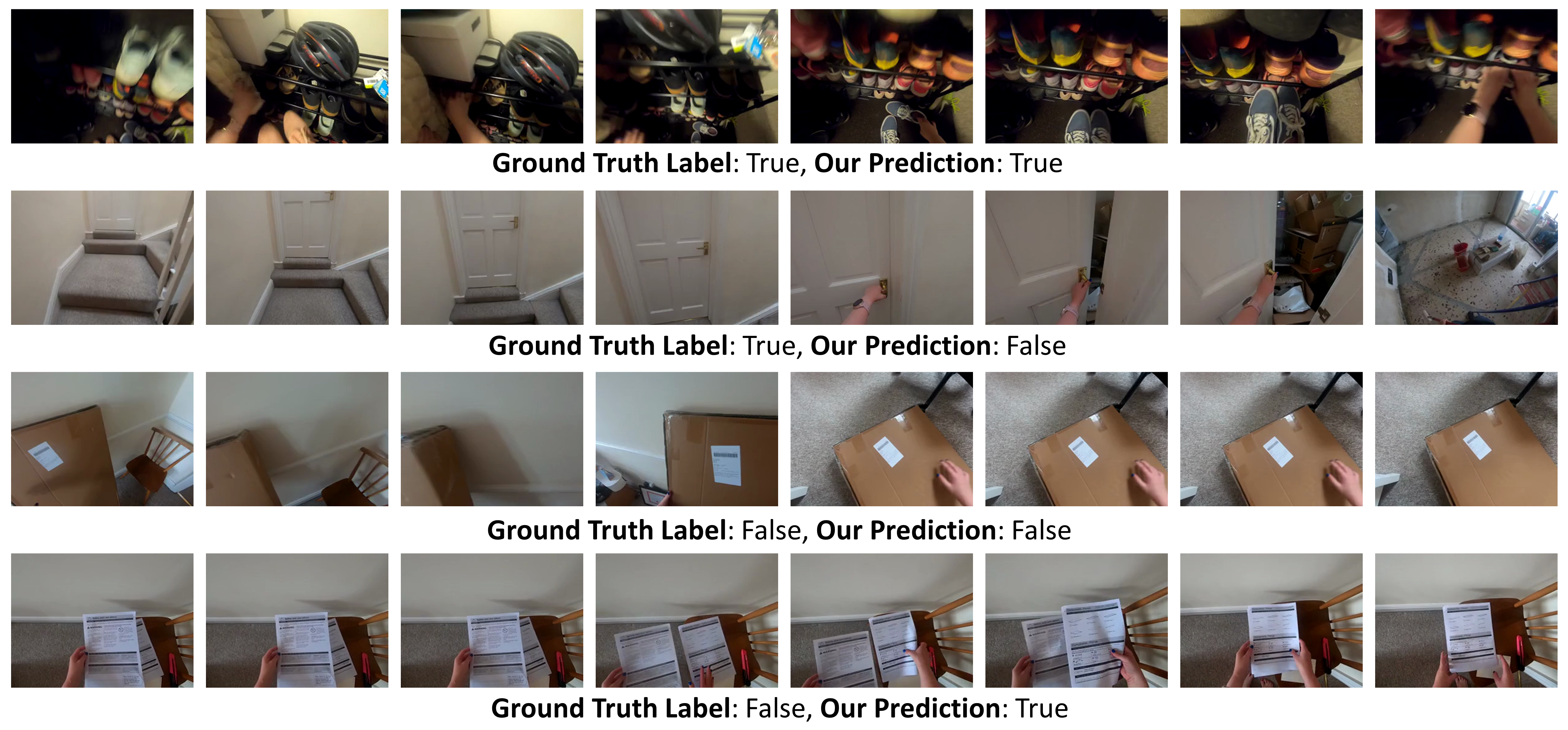}
    \caption{We present several videos of the validation set of the Object State Change Classification task with their ground truth labels and our predictions. Here, we show video frames at 1 fps.}
    \label{fig:example}
\end{figure*}

Figure~\ref{fig:example} shows several videos of the validation set of the Object State Change Classification task with their ground truth labels and our predictions. Specifically, we present one true positive, one false positive, one true negative, and one false negative example. Here, we show video frames at 1 fps. 

\section{Conclusion}

In this report, we describe our submission (``TarHeels") in the Ego4D: Object State Change Classification Challenge. Leveraging the Divided Space-Time attention, we achieve the second-best result on this task. In the future, we will explore more about the egocentric aspects of the videos, and try to improve our model based on the intrinsic properties of egocentric videos. We would also investigate architectures that utilize inductive bias of visual domains, for example, multi-scale video models.

{\small
\bibliographystyle{ieee_fullname}
\bibliography{egbib}
}

\end{document}